\documentclass{article}

\usepackage{graphicx} 
\usepackage{hyperref}
\usepackage{multicol}
\usepackage{url}
\usepackage{makecell}
\usepackage{authblk}
\usepackage[utf8]{inputenc} 
\usepackage[T1]{fontenc}
\usepackage{lmodern}

\usepackage{arxiv}

\usepackage[
backend=biber,
maxbibnames=3,
minbibnames=3,
style=numeric,
]{biblatex}
\AtNextBibliography{\small}

\newbox{\myorcidaffilbox}
\sbox{\myorcidaffilbox}{\large\includegraphics[scale=0.005]{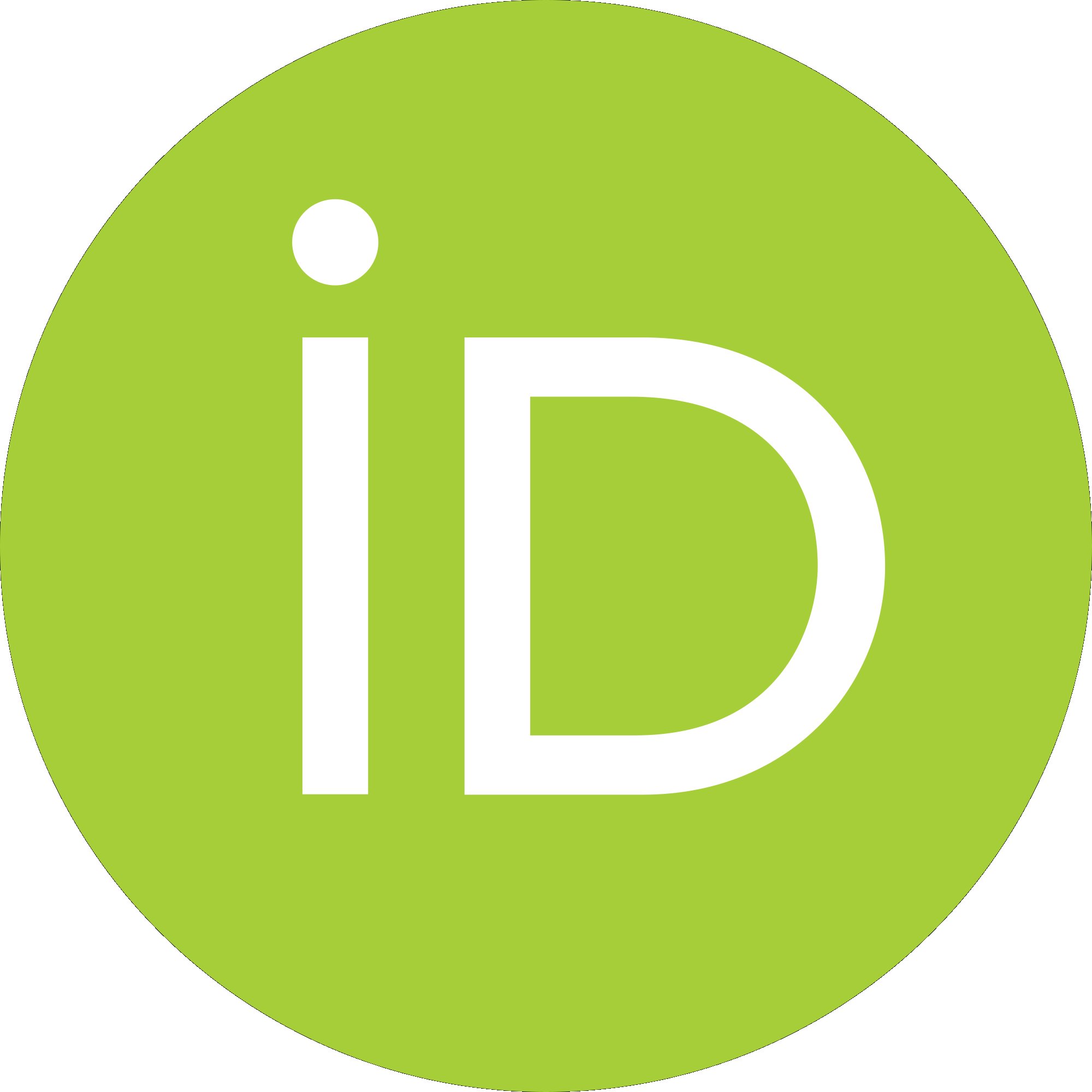}}
\newcommand{\orcidaffil}[1]{%
	\href{https://orcid.org/#1}{\usebox{\myorcidaffilbox}}}
    
\title{Open Llama2 Model for the Lithuanian Language}
\author{Artūras Nakvosas\orcidaffil{0009-0007-7391-5454}, Povilas Daniušis\orcidaffil{0000-0001-5977-827X}, Vytas Mulevičius\orcidaffil{0009-0003-9952-4153}}
\affil{Neurotechnology, Laisvės pr. 125A, LT-06118, Vilnius, Lithuania}

\begin{document}

\maketitle

\begin{abstract}

In this paper, we propose and describe the first open Llama2 large language models (LLMs) for the Lithuanian language, including an accompanying question/answer (Q/A) dataset and translations of popular LLM benchmarks.
We provide a brief review of open regional LLMs and detailed information on the proposed LLMs and their training process. We also conduct an empirical evaluation, comparing the perplexities of the proposed LLMs with those of other modern open LLMs. In addition, benchmarking the proposed LLMs against language understanding tasks reveals that high-quality pretraining datasets may be essential 
for achieving models that perform efficiently on these benchmarks.
The full realisations of the described LLMs are available in the accompanying open repository~\url{https://huggingface.co/neurotechnology}.

\end{abstract}

\section{Introduction}

Large language models (LLMs), relying on Transformer architecture~\cite{NIPS2017_3f5ee243} have shown remarkable effectiveness 
in many natural language processing (NLP) tasks~\cite{minaee2024large,naveed2024comprehensive}. This has primarily been fuelled by increasingly large model parameterisations and training datasets, which are deemed essential according to neural scaling laws~\cite{Hernandez2022ScalingLA}. On the other hand, with the consistent advancement of computational linguistics and NLP, there were recently released open LLMs with performance characteristics comparable with their state-of-the-art (SOTA) commercial counterparts~\cite{touvron2023llama},~\cite{jiang2023mistral},~\cite{jiang2024mixtral},~\cite{almazrouei2023falcon},~\cite{glorioso2024zamba}. Although these open models are useful for further fine-tuning for various downstream problems, training of LLMs usually requires both massive datasets and considerable computational resources.


In the context of the current SOTA, open and commercial LLMs are usually trained on largely English texts~\cite{touvron2023llama}, which results in the lack of performance for less common languages. In addition, commercial LLMs as a rule are not fully accessible (e.g. they are exposed only via APIs, which do not include the model's parameters nor its intermediate representations).
Consequently, there have been multiple recent attempts to achieve efficient open LLMs, tailored for various regional languages other than English (e.g., Section~\ref{sec:related_work}, Table~\ref{table:open_source_llms}). Besides an improved performance for corresponding regional languages, compared to their predecessor LLMs, such open models also are potentially useful for research, as their internal mechanism is fully transparent, and there are related applications, both inside and outside the scope~of NLP~\cite{minaee2024large}.

This article describes Neurotechnology's~\footnote{http://www.neurotechnology.com} contribution to the direction of regional LLM research, encompassing

\begin{itemize}
\item Llama2-based~\cite{touvron2023llama} $7$ and $13$ billion parameter LLMs for the Lithuanian language, and their empirical evaluation,
\item a new dataset, consisting of $13,848$ Q/A pairs primarily about Lithuania and Lithuanian history (in the Lithuanian language) ~\cite{ltqav1},
\item translations of popular LLM benchmarks to the Lithuanian language,
\item open repository, containing all the mentioned components.
\end{itemize}

\noindent  We will structure our paper starting with a short review of the related work in Section~\ref{sec:related_work}. Section~\ref{sec:model} describes our models based on Llama2 architecture and their empirical evaluation. Finally, the conclusive Section~\ref{sec:conclusions} summarises the conducted research from different perspectives.

\section{Related work}
\label{sec:related_work}

\noindent \textbf{Llama2 model.} Transformer-based Llama2 is available in different parameter sizes (e.g. $7B$, $13B$,  and $70B$ parameters), and modifications (e.g. it also includes Llama2-chat version, which is optimised for dialogue use cases). 
The model is first pretrained using a $2$ trillion token set, collected from public sources, and utilising a self-supervised autoregressive approach with cross-entropy loss. Afterwards, it is fine-tuned using publicly available instruction datasets, augmented with human-annotated data, and Reinforcement Learning
with Human Feedback (RLHF) methodologies~\cite{touvron2023llama}.

This model can support the maximum context length of $4096$ tokens. According to benchmarks, Llama2 generally performs on par with many open alternatives (e.g. Falcon~\cite{almazrouei2023falcon}, Mistral~\cite{jiang2023mistral}, Mixtral ~\cite{jiang2024mixtral}, PaLM~ \cite{chowdhery2022palm}, etc.). As is common with large foundational models, it can be further successfully tuned for various downstream tasks, including regional language modelling.

\noindent \textbf{LLMs for regional languages.} Table~\ref{table:open_source_llms} summarises LLMs tailored for common European languages, reflecting the recent contributions from the research and engineering community working in this direction. We include only those regional LLMs, that meet the following criteria:

\begin{itemize}
\item The model should be published in an open repository (e.g. Hugging Face\footnote{https://huggingface.co/}),
\item It should contain at least a minimal description (architecture, training data, and other details).
\end{itemize}

According to Table~\ref{table:open_source_llms}, open LLMs are released for the majority of common European languages.
Table~\ref{table:open_source_llms} shows that Llama2~\cite{touvron2023llama} and Mistral~\cite{jiang2023mistral} are the leading architectures for open LLMs for regional European languages, and $7$ billion parameter models are the most common. Table~\ref{table:open_source_llms} also reveals that full-parameter training is conducted in the majority of cases ($19$ cases from $20$), instead of the parameter-efficient fine-tuning (PEFT) based approach. However, in some instances ($2$ cases from $20$) regional LLMs were trained using PEFT methods (e.g., LoRA~\cite{hu2021lora}, MoRA~\cite{jiang2024mora}), which may result in less accurate models compared to full-parameter training, although with the lower computational costs. In addition, quite often only the model itself is published ($11$ out of $20$ cases), without an accompanying citable document (e.g. technical report/peer-reviewed publication), or training and evaluation datasets. In our opinion, the lack of accompanying scientific documentation limits the potential usefulness of the released regional LLMs in various important aspects, including their reproducibility, empirical performance assessment, and establishing a connection to the existing related results.

\begin{table}[!h]
\begin{center}
\begin{tabular}{|l|c|c|c|c|c|}
\hline
\textbf{Language} & \textbf{Architecture} & \textbf{Size} & \textbf{Ref.}  & \textbf{F/P} & \textbf{Doc.} \\
\hline
 Bulgarian & Mistral & 7B & ~\cite{insait_bggpt_2024} &  F & No \\ \hline
 Danish & Mistral & 7B & ~\cite{munin7b_alpha,heidrun_mistral_7b_chat} &  F  & No \\ \hline
 Dutch & Mistral & 7B & ~\cite{rijgersberg_geitje}  & F & No \\ \hline
 French-English & Llama & 1.3B & ~\cite{faysse2024croissantllm} & F & Yes \\ \hline 
 German & Llama2 & 7B,13B & ~\cite{LeoLM} & F & No \\ \hline 
 Greek & Mistral & 7B & ~\cite{SPAHE2024Meltemi} & F & No \\ \hline 
 Hungarian-English & Llama2 & 7B &~\cite{csaki2024sambalingo} & F  & Yes \\ \hline 
 Finnish and other & Llama2  & 7B-33B & \cite{viking13b} &  F & No \\ \hline 
 Icelandic & RoBERTa & & \cite{snæbjarnarson2022warmstartcleancrawled} & F & Yes \\ \hline  
 Italian & Llama2 & 7B,13B & ~\cite{bacciu2023fauno} & P & Yes \\ \hline
 Lithuanian & Llama2  & 7B,13B &  \textbf{Ours} & F & Yes  \\ \hline 
 Norwegian & Mistral  & 7B & \cite{normistral_7b_warm,} & F & No \\ \hline  
 \begin{tabular}{@{}l@{}}Serbian, Bosnian,\\ Croatian \end{tabular} & Mistral & 7B & \cite{yugogpt}  & F  & No \\ \hline  
 Spanish & Falcon & 7B  & \cite{aguila7b} & F & No \\ \hline 
 Swedish & GPT-SW3 &  126M-40B & \cite{ekgren2023gptsw3} &  F & Yes \\ \hline  
 Slovenian & RoBERTa &   & \cite{ulcar2021sloberta} & F & Yes\\ \hline   
 Polish & Mistral  & 7B  & \cite{bielik_7b_instruct_v0.1,} & F & No \\ \hline 
 Ukrainian & Mistral & 7B & \cite{boros-chivereanu-dumitrescu-purcaru-2024-llm-uk} & F & Yes \\ \hline
 Portuguese & Phi-2B & 1.3B-7B &  \cite{garcia2024introducingbodefinetunedlarge} & P  & Yes \\ \hline 
Romanian  & Llama2 & 7B  & \cite{masala2024openllmro}  & F & Yes \\ \hline 
\end{tabular}
\caption{Open LLM models for regional European languages. The F/P column denotes whether the model was full-parameter trained (F), or trained via PEFT (P). Doc. column shows whether the corresponding model has an accompanying publication.}
\label{table:open_source_llms}
\end{center}
\end{table}

\section{Proposed open LLMs and their evaluations}
\label{sec:model}

\begin{figure}[h]
\centering
\includegraphics[width=0.6\textwidth]{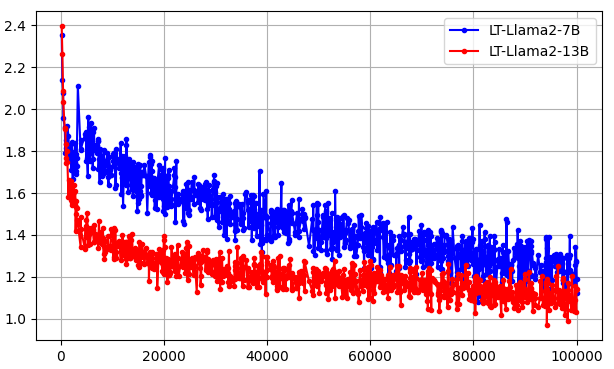}
\caption{Losses (y-axis) vs training steps (x-axis) during the model's pretraining.}
\label{fig:training_loss}
\end{figure}

\begin{figure}[h]
\centering
\includegraphics[scale=0.5]{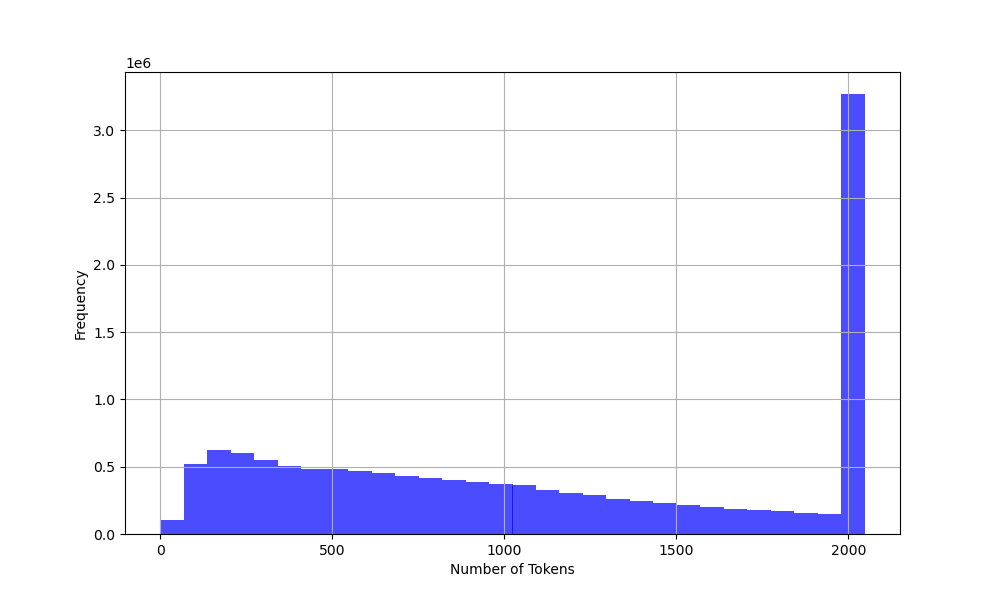}
\caption{Distribution of the record length of the Lithuanian component of the CulturaX dataset (in tokens).}
\label{fig:culturax_record_len_distribution}
\end{figure}

\noindent \textbf{Proposed open LLMs and their training details.}  The proposed models (including tokenizers) are trained from Llama2 $7$~\footnote{https://huggingface.co/meta-llama/Llama-2-7b} and $13$~\footnote{https://huggingface.co/meta-llama/Llama-2-13b} billion parameter checkpoints (further denoted by Llama2-7B and Llama2-13B, correspondingly).
The training consists of two phases: the first one is standard autoregressive pretraining on the Lithuanian component of the CulturaX dataset~\cite{nguyen2023culturax}, and the second one is fine-tuning on the  Alpaca~\cite{alpaca} dataset, which has been translated into Lithuanian using ChatGPT (gpt-4-1106-preview) and~\cite{ltqav1} dataset. We train the full model without using PEFT. Figure~\ref{fig:culturax_distribution} shows the source distribution of the Lithuanian component of the CulturaX dataset, and Figure~\ref{fig:culturax_record_len_distribution} shows the record length distribution in tokens. We use $2048$ token context length during model training for both models. The models are trained on 8xH100 GPUs. Figure~\ref{fig:training_loss} shows loss during the model pretraining process.  The training details for both LLMs are provided in Table~\ref{table:training_details} and the fine-tuning was conducted with the same parameters, except the learning rate, which was set to $0.00001$. 
The download links for all the proposed LLMs are provided in Table~\ref{table:download_links}.



\begin{table}[!h]
\begin{center}
\begin{tabular}{|l|c|c|}
\hline
\textbf{Learning parameter} & \textbf{Llama2-7B} & \textbf{Llama2-13B}  \\
\hline
 Number of epochs & $1$ & $1$ \\ \hline
 Learning rate &  $0.0002$ & $0.00004$ \\ \hline
 Warmup ratio &  $0.05$ & $0.05$ \\ \hline
 Weight decay &  $0.07$ & $0.05$ \\ \hline 
 Per-device batch size & $8$ &  $4$ \\ \hline
 Gradient accumulation steps & $2$ & $4$\\ \hline 
\begin{tabular}{@{}l@{}} Duration in seconds for a \\ single H100 GPU \end{tabular}  & $1722.0$  & $2980.5$  \\ \hline
 Total number of tokens & \multicolumn{2}{c|}{14761219995} \\ \hline
 Records in dataset & \multicolumn{2}{c|}{13339785} \\ \hline
 Mean number of tokens per record & \multicolumn{2}{c|}{1106.5560} \\ \hline
 Standard deviation of tokens per record & \multicolumn{2}{c|}{697.0089} \\ \hline
\end{tabular}
\caption{Training details.}
\label{table:training_details}
\end{center}
\end{table}

\noindent \textbf{Proposed open Q/A dataset ~\cite{ltqav1}.} This dataset was constructed from the ChatGPT~\cite{openai2024gpt4} summarisations of a set of Lithuanian Wikipedia pages, and represents a collection of $13,848$ Q/A pairs primarily about Lithuania and Lithuanian history. It was not used during pretraining process. Table~\ref{table:qa_examples} showcases a set of examples from this dataset. 

\begin{table}[h!]
\centering
\begin{tabular}{|m{7cm}|m{8cm}|}
\hline
\textbf{Question} & \textbf{Answer} \\ \hline
Koks yra Vilniaus miesto statusas Lietuvoje?	& Vilnius yra Lietuvos sostinė.\\ \hline
Kur yra Gedimino pilis? & Gedimino pilis yra Vilniuje, ant Gedimino kalno. \\ \hline
Kas buvo vadinamas „Lito tėvu“?& Vladas Jurgutis buvo vadinamas „Lito tėvu“, nes jam buvo patikėta spręsti visus naujos valiutos įvedimo niuansus.\\ \hline
Kokios upės teka per Vilnių? & Per Vilnių teka Neris ir Vilnia. \\ \hline
Kada buvo įkurtas Vilniaus universitetas?	& Vilniaus universitetas buvo įkurtas 1579 metais, Vilniuje, po Lietuvos didžiojo kunigaikščio Stepono Batoro privilegijos suteikimo jėzuitų ordino kolegijai. \\ \hline
Kada ir kur įvyko Žalgirio mūšis? & Žalgirio mūšis įvyko 1410 m. liepos 15 d. netoli Tanenbergo ir Griunvaldo (Žalgirio) kaimelių, dabartinės Lenkijos teritorijoje, į pietvakarius nuo Olštyno.  \\ \hline
\end{tabular}
\caption{Examples from the accompanying Q/A dataset.}
\label{table:qa_examples}
\end{table}

\subsection{Perplexity evaluation benchmarks}
We further conduct an empirical evaluation of the proposed Lithuanian LLMs using an open Lithuanian language dataset \cite{ltqav1}. We analyse LLMs by examining their perplexity, which is defined as

\begin{equation}
\label{eq:perplexity}
P(W) := \exp\left( - \frac{1}{N} \sum_{i=1}^{N} \log p(w_i \mid w_1, w_2, \ldots, w_{i-1}) \right),
\end{equation}
where
\begin{itemize}
  \item \( W = w_1, w_2, \ldots, w_N \) is the sequence of tokens,
  \item \( N \) is the number of tokens in the sequence,
  \item \( p(w_i \mid w_1, w_2, \ldots, w_{i-1}) \) is the conditional probability of the token \( w_i \) given the previous tokens \( w_1, w_2, \ldots, w_{i-1} \).
\end{itemize}

It can be interpreted as the model's ability to predict the next token. From the definition (Eq. ~\ref{eq:perplexity}), the lower perplexity values indicate better performance, and for any input sequence $W$, $P(W) \geq 1$. The selection of input perplexity was motivated by~\cite{gonen-etal-2023-demystifying},
where the authors reveal that for a wide range of tasks, the lower the perplexity of the prompt is, the better the prompt can perform the task. We analyse the average perplexity (averaged over all Q/A concatenations) as the performance measure.
We compare empirical perplexities of five LLMs, further denoted as Llama2-7B, LT-Llama2-7B, Llama2-13B, LT-Llama2-13B, and Llama3-8B.
Llama2-7B and Llama2-13B correspond to the default Llama2 models with $7$ and $13$ billion parameters, and Llama3-8B denotes the default $8$ billion parameter Llama3 LLM~\cite{Llama3}. This model was trained on a multilingual dataset that also includes a significant proportion of Lithuanian data. The proposed $7$ and $13$ billion parameter Llama2 LLMs, trained for the Lithuanian language are denoted as LT-Llama2-7B and Llama2-13B. Table~\ref{table:open_source_llm_evaluation} shows the average perplexities of the aforementioned LLMs. According to it, the proposed LT-Llama2-7B and LT-Llama2-13B models exhibit significantly lower average perplexity values compared to the Llama3-8B.

\begin{table}[!h]
\begin{center}
\begin{tabular}{|l|c|}
\hline
\textbf{Model} & \textbf{Average perplexity}  \\
\hline
 Llama2-7B & 17.4613 \\ \hline
 LT-Llama2-7B  & 3.8096 \\ \hline 
 Llama2-13B & 13.8849 \\ \hline
 LT-Llama2-13B & 3.4520 \\ \hline 
 Llama3-8B &  5.9795 \\ \hline
\end{tabular}
\caption{Average perplexities.}
\label{table:open_source_llm_evaluation}
\end{center}
\end{table}

\begin{figure}[h]
\centering
\includegraphics[width=10.0cm]{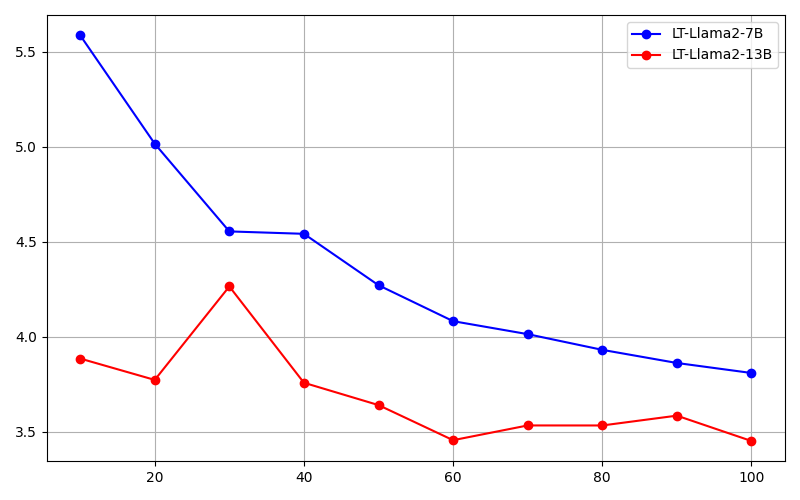}
\caption{Percentage of the Lithuanian component of the CulturaX dataset used in the pretraining (x-axis) vs. corresponding average perplexity (y-axis).}
\label{fig:culturax_perplexity}
\end{figure}

\noindent \textbf{Proportion of pretraining data versus average perplexity.} In this experiment with the same~\cite{ltqav1} dataset, we investigate the association of the average perplexity measured on~\cite{ltqav1}, and the percentage of the data from CulturaX Lithuanian component, used in the pretraining process. We conduct this experiment using both LT-Llama2-7B and LT-Llama2-13B models,  saving the model's parameters every $10\%$ of the total number of iterations during a pretraining epoch.
Figure~\ref{fig:culturax_perplexity} reveals that with the inclusion of additional training data, the perplexity tends to decrease, although, in the end, increasing saturation is visible in both cases. According to Table~\ref{table:open_source_llm_evaluation}, the initial perplexities are $17.4613$ and $13.8849$, correspondingly. Interestingly, pretraining even with $10\%$ of data results in models with smaller perplexity (measured on \cite{ltqav1}) than that of the Llama3 model. The perplexities of the LT-Llama2-13B are also uniformly lower compared to those of the smaller LT-Llama2-7B model.

\subsection{Language understanding benchmarks}

\begin{figure}[h]
\centering
\includegraphics[width=15.5cm]{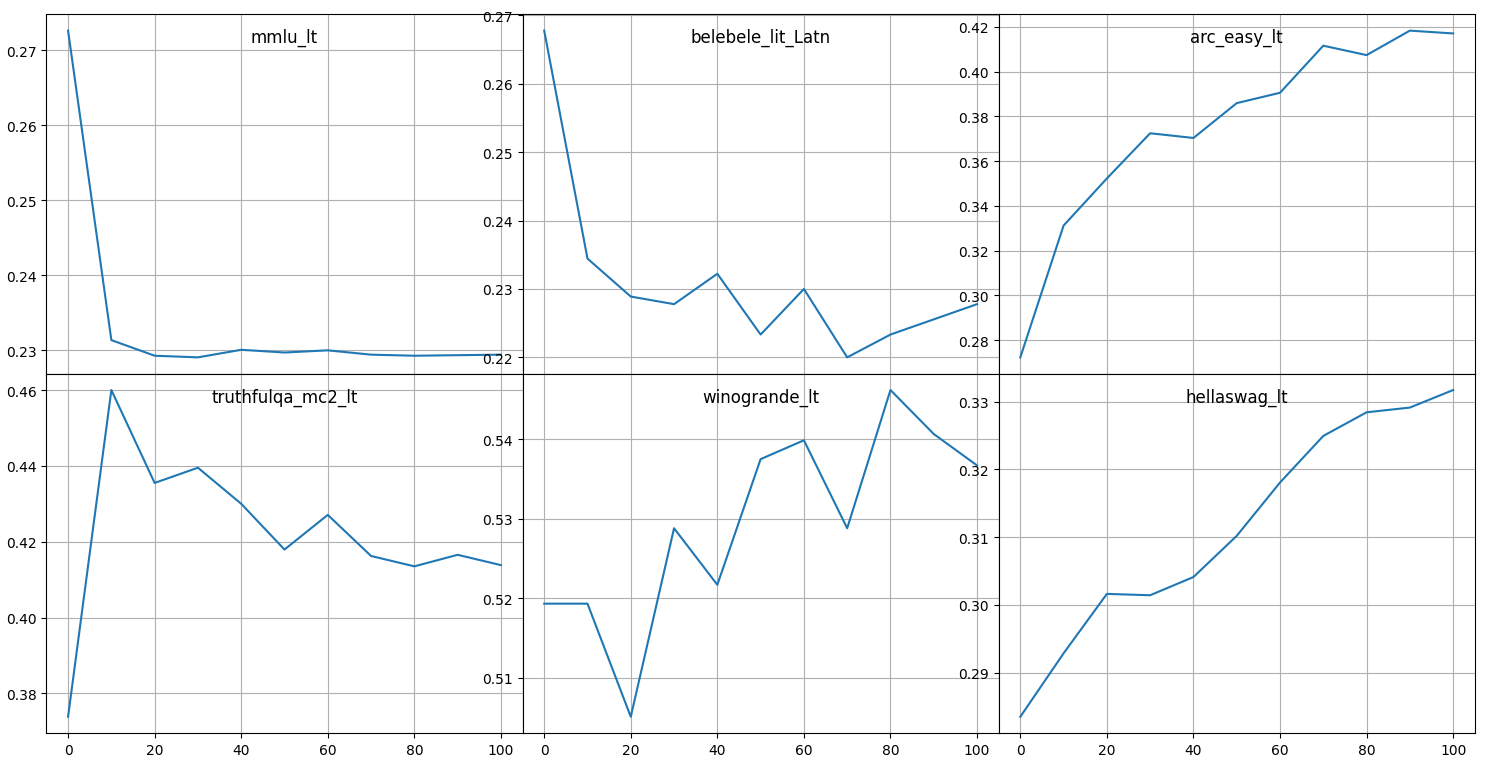}
\caption{Accuracies (y-axis) of LMEH benchmarks for LT-Llama2-7B model, pretrained with different proportions of Lithuanian component of CulturaX dataset (x-axis). The MMLU benchmarks are summarized in mmlu\_lt.}
\label{fig:benchmark_7b}
\end{figure}

\begin{figure}[h]
\centering
\includegraphics[width=15.5cm]{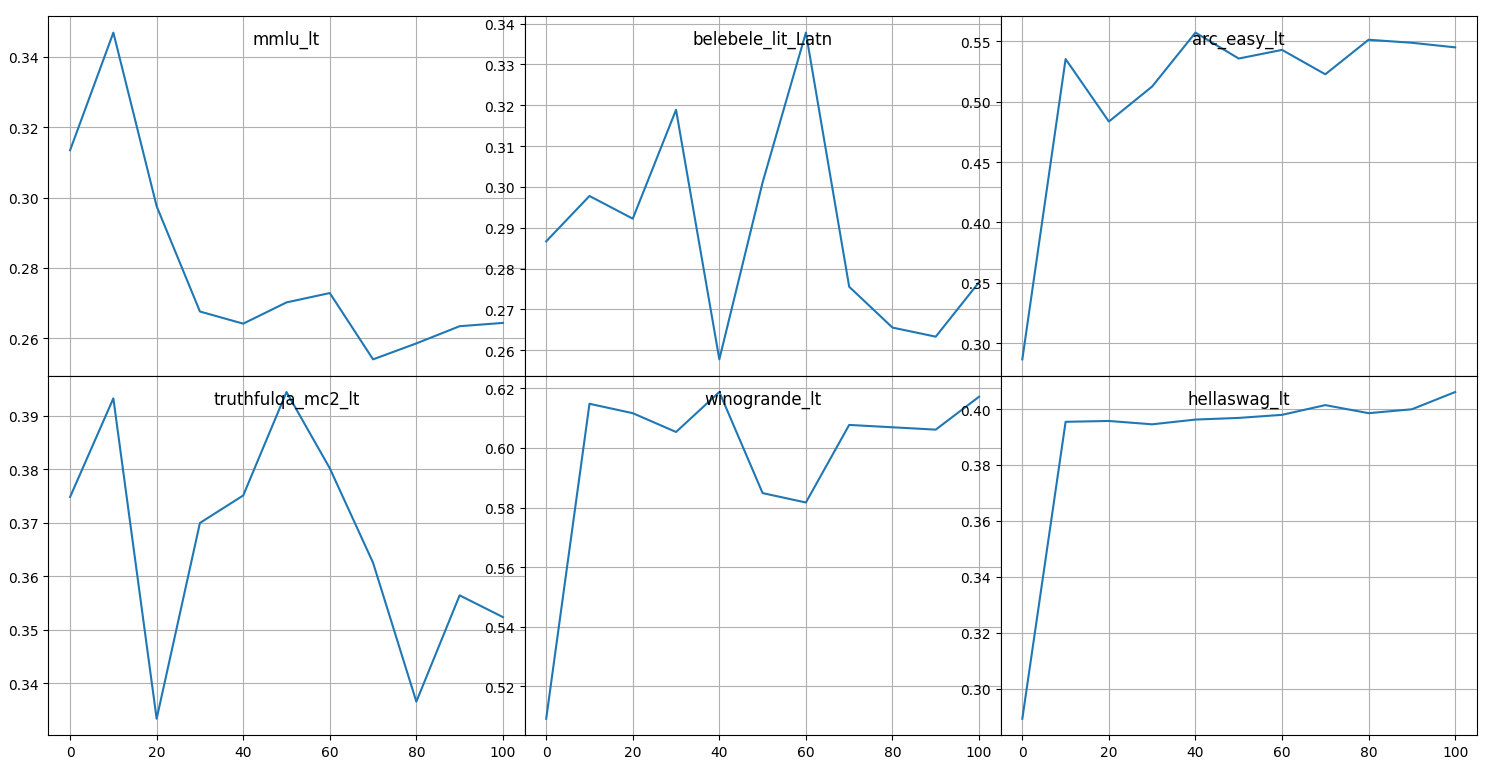}
\caption{Accuracies (y-axis) of LMEH benchmarks for LT-Llama2-13B model, pretrained with different proportions of Lithuanian component of CulturaX dataset (x-axis). The MMLU benchmarks are summarized in mmlu\_lt.}
\label{fig:benchmark_13b}
\end{figure}

\begin{figure}[h]
\centering
\includegraphics[width=1.0\textwidth,height=1.1\textwidth]{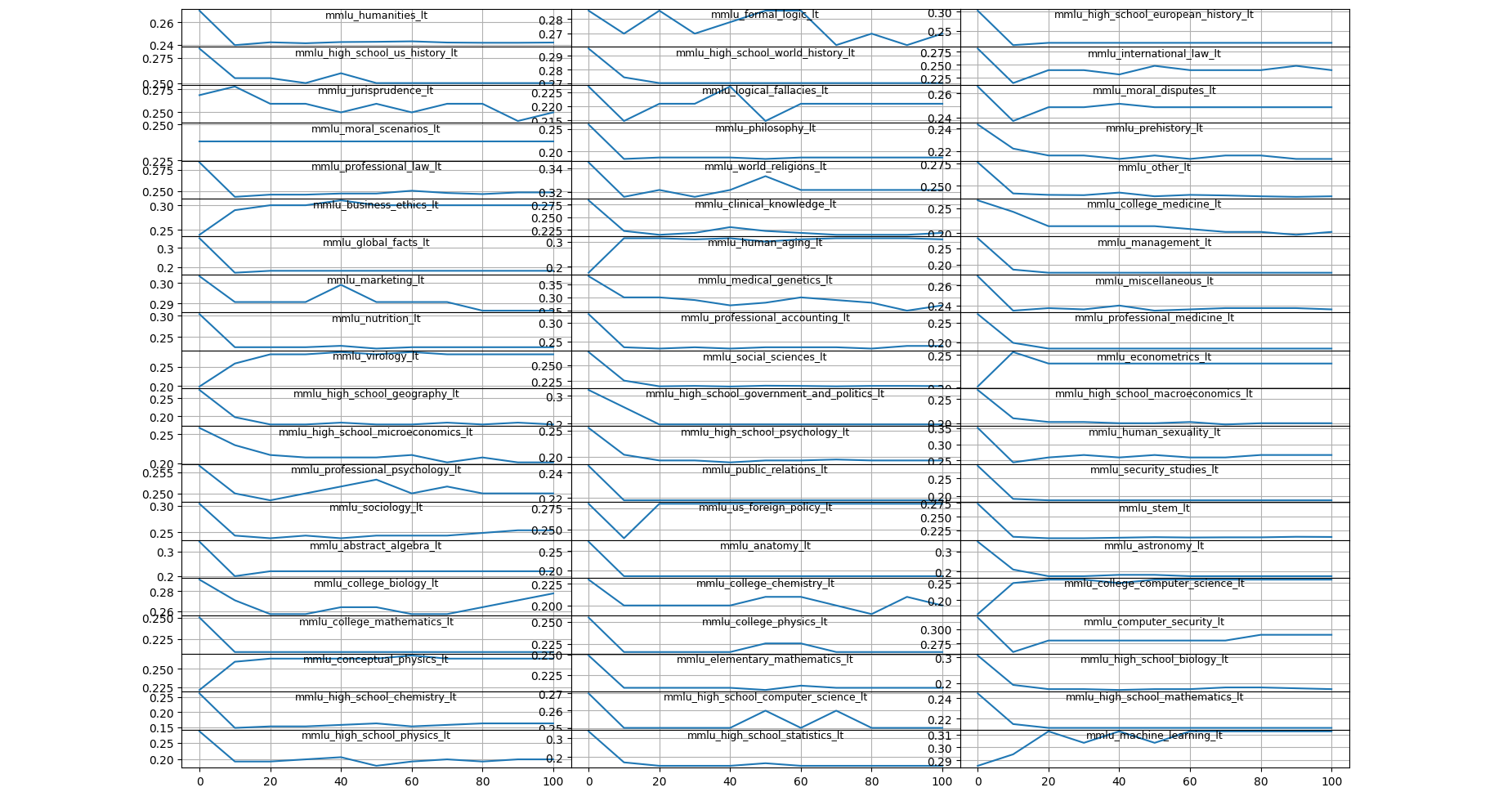}
\caption{Accuracies (y-axis) of individual MMLU benchmarks for LT-Llama2-7B model, pretrained with different proportions of Lithuanian component of CulturaX dataset (x-axis).} 
\label{fig:mmlu_7b}
\end{figure}

\begin{figure}[h]
\centering
\includegraphics[width=1.0\textwidth,height=1.1\textwidth]{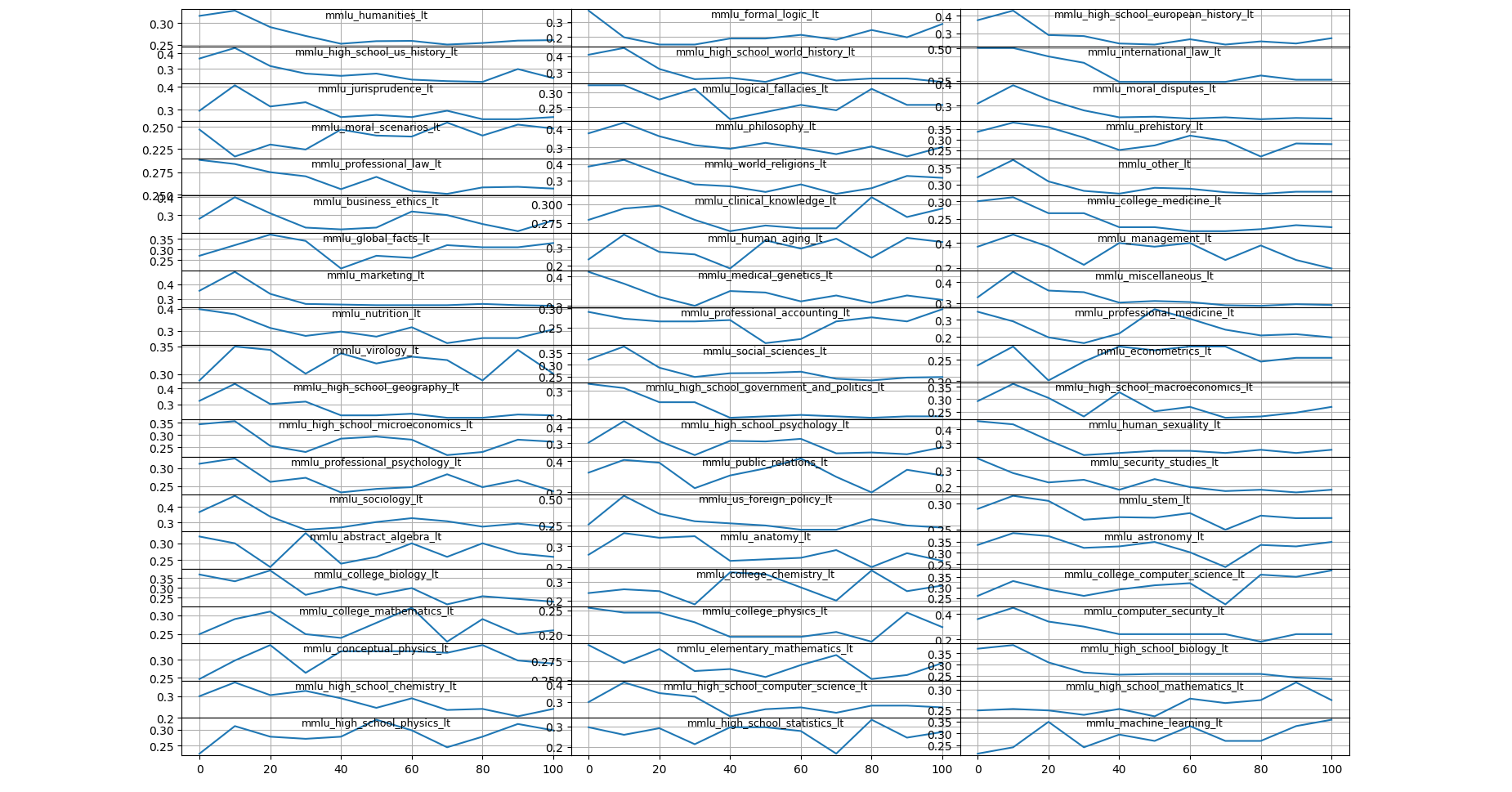}
\caption{Accuracies (y-axis) of individual MMLU benchmarks for LT-Llama2-13B model, pretrained with different proportions of Lithuanian component of CulturaX dataset (x-axis).} 
\label{fig:mmlu_13b}
\end{figure}

In this experiment, we evaluate the proposed LLMs using LM evaluation harness (LMEH) language understanding benchmarks~\cite{eval-harness} translated into Lithuanian language (see Table~\ref{table:download_links} for the download links). These benchmarks are created to assess LLMs across a wide range of evaluation tasks including Massive Multitask Language Understanding (MMLU) set~\cite{hendrycks2021measuringmassivemultitasklanguage}, primarily covering diverse academic disciplines.  Figure~\ref{fig:benchmark_7b} and Figure~\ref{fig:benchmark_13b}  showcases the accuracies for a sequence of checkpoints, which correspond to the percentage of the pretraining data from CulturaX Lithuanian component, starting with $0\%$ (which corresponds to the initial Llama2-7B), with the step of $10\%$. Similarly, Figure~\ref{fig:mmlu_7b} and Figure~\ref{fig:mmlu_13b} provide information about individual benchmarks from the MMLU set.

\begin{figure}[h]
\centering
\includegraphics[angle=270,scale=0.34]{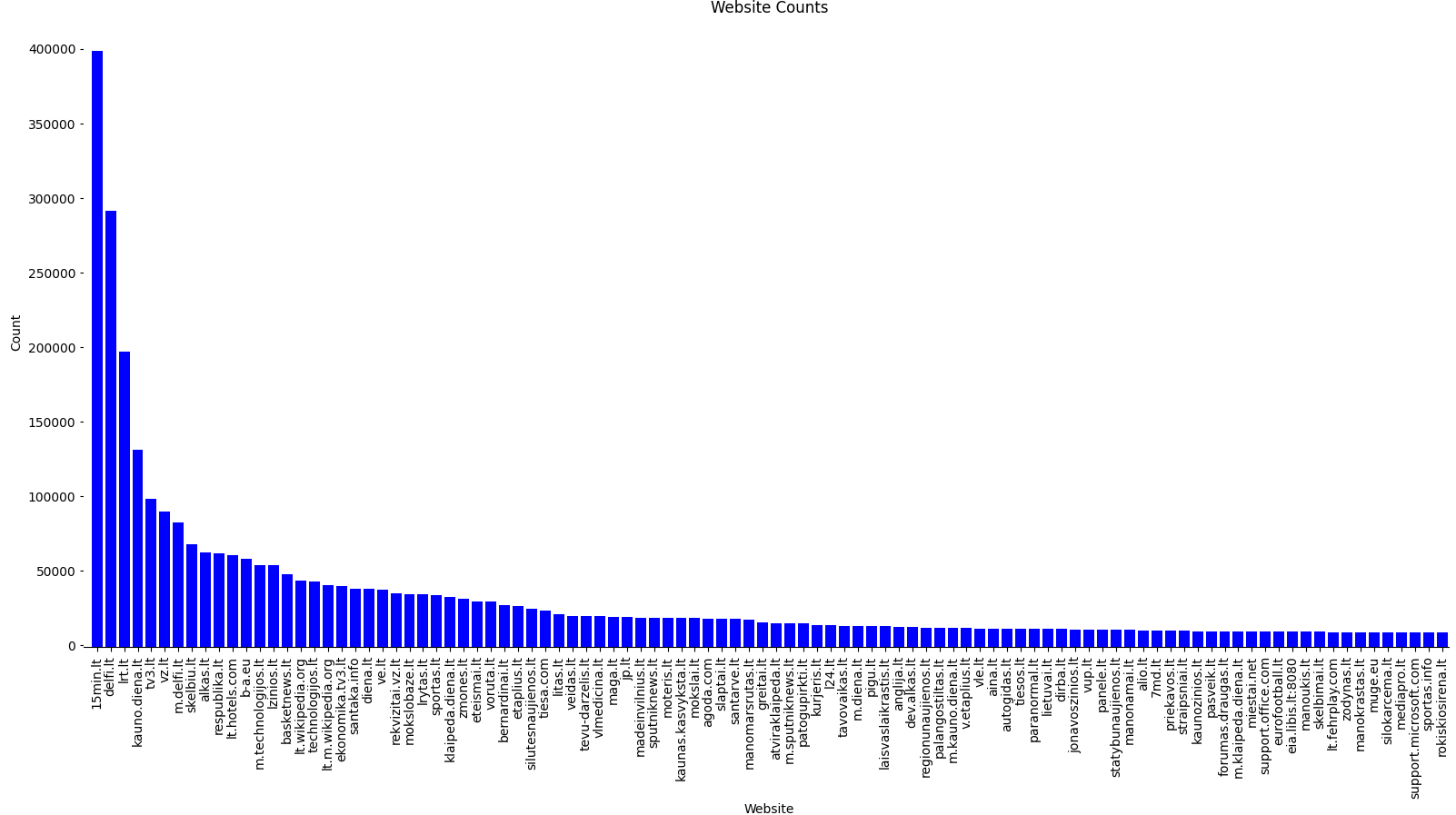}
\caption{Source distribution of the Lithuanian component of the CulturaX dataset.}
\label{fig:culturax_distribution}
\end{figure}

Although for some tasks (e.g., \verb|arc_easy_lt|, \verb|hellaswag_lt|, \verb|winogrande_lt|) we see consistent improvement throughout the entire pretraining process, this benchmark surprisingly reveals that in most cases of MMLU (see Figure~\ref{fig:mmlu_7b} and Figure~\ref{fig:mmlu_13b}), there is no improvement compared to the initial model. We hypothesise that this may be because the Lithuanian component of CulturaX is almost exclusively collected through web crawling of common websites (see Figure~\ref{fig:culturax_distribution}), which does not include data that is relevant to those specific tasks. Therefore, the extensions of regional components of CulturaX with high-quality data may improve LLMs, tailored for the corresponding regional languages.

\section{Conclusions}
\label{sec:conclusions}

We presented the first Llama2-based open LLMs tailored especially for the Lithuanian language. Our model is released with the accompanying QA dataset~\cite{ltqav1} and translated standard LLM benchmarks. 

The motivation for our research was to achieve Lithuanian LLMs, which were either nonexistent (e.g., Llama2~\cite{touvron2023llama}) or quite weak (e.g., Llama3~\cite{Llama3}). Although the latest open multilingual models (e.g., Llama3.1~\cite{Llama3}, Gemma2~\cite{gemmateam2024gemma2improvingopen}), released during our research, have a strong Lithuanian component, we trained our model based on Llama2 to investigate whether an efficient Lithuanian LLM can be trained from an LLM without any Lithuanian component.

We also conducted an overview of the existing regional LLMs. It shows that most regional models follow Llama2 or Mistral architecture. In addition, some authors do not train a full parameter set but instead rely on PEFT approaches (e.g. ~\cite{hu2021lora}), which are less computationally demanding, but also potentially less efficient in performance. On the other hand, PEFT methods partially allow to retain the original parameter structure, and thereby they may be beneficial for achieving more efficient regional LLMs from the perspective of language understanding benchmarks, such as MMLU. Table~\ref{table:open_source_llms} also reveals a lack of scientific documentation of the published open regional LLMs. 
We also conducted benchmarks of our model with~\cite{ltqav1}, evaluating average perplexities during its pretraining on different proportions CulturaX dataset. The results show that in these benchmarks our model surpasses the default Llama3.
In addition, we evaluated the proposed LLMs with LMEH, which includes a conceptually diverse set of language model benchmarks. The results of these experiments hint that the Lithuanian component of CulturaX may not be sufficiently rich for modern LLM architectures. We also translated these benchmarks into Lithuanian and published them in an open repository (see Table~\ref{table:download_links}), contributing to the standardisation of Lithuanian language model evaluation.

In the context of regional LLMs, the proposed models open further research perspectives not only for NLP, but also for other directions since LLM representations are potentially useful in various scenarios (e.g. sentiment analysis~\cite{zhang-etal-2024-sentiment}, robotics~\cite{kim2024surveyintegrationlargelanguage}, causality~\cite{liu2024largelanguagemodelscausal}, and multimodality ~\cite{parekh2024conceptbasedexplainabilityframeworklarge}). Our future work will include fully trained small language models tailored for Baltic languages and English.

\section*{Acknowledgement}

This research was funded by Neurotechnology. We are grateful to Neurotechnology for providing resources and support for this research. We also thank Rasa Kundrotaitė for editing the English language, Ignas Mataitis, and other colleagues for useful remarks and discussions.

\medskip

\printbibliography

\newpage
\section*{Appendix}

\newpage

\begin{table}[!b]
\begin{center}
\begin{tabular}{|l|}
\hline
\textbf{Download links for the proposed LLMs}  \\
\hline
 \href{https://huggingface.co/neurotechnology/Lt-Llama-2-7b-hf}{LT-Llama2-7B (pretrained)} \\ \hline 
\href{https://huggingface.co/neurotechnology/Lt-Llama-2-7b-instruct-hf} {LT-Llama2-7B (pretrained and fine-tuned)} \\ \hline 
 \href{https://huggingface.co/neurotechnology/Lt-Llama-2-13b-hf} {LT-Llama2-13B (pretrained)} \\ \hline 
 \href{https://huggingface.co/neurotechnology/Lt-Llama-2-13b-instruct-hf}{LT-Llama2-13B (pretrained and fine-tuned)} \\ \hline
\textbf{Download links for the translated data sets} \\
\hline 
\href{https://huggingface.co/datasets/neurotechnology/lt_arc}{LT-Arc~\cite{lai2023okapiinstructiontunedlargelanguage}} \\ \hline

\href{https://huggingface.co/datasets/neurotechnology/lt_winogrande}{LT-Winogrande~\cite{sakaguchi2019winograndeadversarialwinogradschema}} \\ \hline

\href{https://huggingface.co/datasets/neurotechnology/lt_mmlu}{LT-MMLU~\cite{hendrycks2021measuringmassivemultitasklanguage}} \\ \hline

\href{https://huggingface.co/datasets/neurotechnology/lt_thruthful_qa}{LT-Truthful-qa~\cite{lai2023okapiinstructiontunedlargelanguage}} \\ \hline

\href{https://huggingface.co/datasets/neurotechnology/lt_hellaswag}{LT-Hellaswag~\cite{zellers2019hellaswag}} \\ \hline

\href{https://huggingface.co/datasets/neurotechnology/lt_gsm8k}{LT-GSM8K~\cite{cobbe2021trainingverifierssolvemath}} \\ \hline

\end{tabular}
\caption{Download links for proposed LLMs and data.}
\label{table:download_links}
\end{center}
\end{table}

\end{document}